\documentclass{article}

\PassOptionsToPackage{numbers,compress}{natbib}
 \usepackage[preprint]{neurips_2026}

\usepackage[utf8]{inputenc} 
\usepackage[T1]{fontenc}    
\usepackage[hidelinks]{hyperref}       
\usepackage{url}            
\usepackage{booktabs}       
\usepackage{array}          
\usepackage{longtable}      
\usepackage{seqsplit}       
\usepackage{amsfonts}       
\usepackage{amssymb}        
\usepackage{nicefrac}       
\usepackage{microtype}      
\usepackage[table]{xcolor}  
\usepackage{amsmath}
\usepackage{graphicx}
\usepackage[most]{tcolorbox}
\tcbuselibrary{listings,breakable}
\usepackage{wrapfig}
\usepackage{enumitem}
\usepackage{caption}

\newtcblisting{markdownbox}[1]{
  enhanced,
  breakable,
  listing only,
  title={#1},
  fonttitle=\bfseries,
  colback=white,
  colframe=black!60,
  colbacktitle=black!8,
  coltitle=black,
  boxrule=0.6pt,
  arc=1.5mm,
  left=2mm,
  right=2mm,
  top=1mm,
  bottom=1mm,
  listing options={
    basicstyle=\ttfamily\scriptsize,
    breaklines=true,
    breakatwhitespace=false,
    columns=fullflexible,
    keepspaces=true,
    showstringspaces=false
  }
}

\newcommand{\toolyes}{\checkmark}
\newcommand{\toolno}{\ensuremath{\times}}
\definecolor{retoolrow}{RGB}{233, 245, 236}
\DeclareRobustCommand{\seqsplit}[1]{%
  \begingroup
  \edef\toolnameexpanded{#1}%
  \expandafter\nolinkurl\expandafter{\toolnameexpanded}%
  \endgroup
}

\newcommand\blfootnote[1]{%
  \begingroup
  \renewcommand\thefootnote{}\footnote{#1}%
  \addtocounter{footnote}{-1}%
  \endgroup
}

\title{
\textsc{ReTool-Video}: Recursive Tool-Using Video Agents with Meta-Augmented Tool Grounding
}

\author{
\textbf{Xiao Liu\textsuperscript{\rm 1}, 
Nayu Liu\textsuperscript{\rm 2}, 
Junnan Zhu\textsuperscript{\rm 3}, 
Ruirui Chen\textsuperscript{\rm 4}, 
Guohui Xiang\textsuperscript{\rm 5}, 
Changjian Wang\textsuperscript{\rm 5}}, 
\\ \textbf{Kaiwen Wei\textsuperscript{\rm 1,$\dagger$}, 
Rongzhen Li\textsuperscript{\rm 5,$\dagger$}, 
Jiang Zhong\textsuperscript{\rm 1}}\\
\textsuperscript{\rm 1}Chongqing University \\
\textsuperscript{\rm 2}Tianjin University \\
\textsuperscript{\rm 3}MAIS, Institute of Automation, Chinese Academy of Sciences \\
\textsuperscript{\rm 4}Institute of High Performance Computing (IHPC),\\ Agency for Science, Technology and Research (A*STAR), Singapore  \\
\textsuperscript{\rm 5}Chongqing National Data AI Research Institute, AI Research Lab \\
\texttt{\{Liu-xiao-\}@outlook.com}, \texttt{\{weikaiwen\}@cqu.edu.cn}
}

\begin{document}
\maketitle
\blfootnote{$\dagger$ means corresponding authors.}

\begin{abstract}
Video understanding requires active evidence seeking, motivating tool-augmented video agents for temporal reasoning, cross-modal understanding, and complex question answering. Existing video agents have improved video reasoning with retrieval, memory, frame inspection, and verifier tools, but they still face two limitations: (1) a coarse tool space that lacks fine-grained operations for compositional reasoning; and (2) a flat action space that forces high-level video intents into primitive executable tool calls. In this paper, we address these challenges with two complementary designs. First, we construct a MetaAug-Video Tool Library (MVTL), an extensible tool library with 134 registered tools, including 26 base tools for general multimodal signal processing and 108 meta tools for filtering, aggregation, reranking, formatting, and other intermediate-result operations. MVTL supports dual-level access to both structured video information and raw modal evidence, enabling diverse video reasoning scenarios. Second, we propose \textsc{ReTool-Video}, a recursive tool-using method that grounds high-level video intents into executable tool chains. In \textsc{ReTool-Video}, matched actions are executed directly, while unmatched intents are delegated to a resolver for parameter repair, tool substitution, or decomposition. This allows abstract actions such as temporal merging, cross-modal verification, or repeated-event aggregation to be progressively translated into concrete multimodal operations at runtime. Experiments on MVBench, MLVU, and Video-MME$_{\mathrm{w/o\ sub.}}$ show that \textsc{ReTool-Video} consistently outperforms strong baselines. Further analysis demonstrates that recursive grounding and fine-grained meta tools improve the stability and effectiveness of complex video understanding. 
\end{abstract}

\section{Introduction}

Video understanding is a fundamental capability for multimodal models in real-world scenarios, where videos contain temporal dynamics, cross-modal signals, and structured information such as audio, subtitles, object-state changes, and event transitions. These properties support tasks such as temporal localization, dense event description, compositional question answering, action counting, state tracking, and cross-modal reasoning~\citep{krishna2017dense,lei2018tvqa,xiao2021next,li2024mvbench,fu2025video}. Recent multimodal models, from GPT-4V to Claude-Sonnet-4.5, have shown strong visual and video understanding abilities~\citep{openai2023gpt4v,anthropic2025claudesonnet45systemcard}, while open-source video-language models improve through alignment, instruction tuning, long-context modeling, sparse memory, and hierarchical retrieval~\citep{maaz2024video,lin2024video,zhang2024long,ma2025drvideo}. However, complex video QA often requires active reasoning beyond single-step perception, including moment localization, evidence inspection, cross-modal verification, observation aggregation, and external tool use. This motivates video agents that integrate planning, execution, memory, retrieval, and verification capabilities~\citep{fan2024videoagent,zhang2025deep,zhang2025thinking,li2026videothinker,lin2026videoseek,rege2026agentic}. Scaling such agents depends on both the tool space that provides multimodal capabilities and the action space that maps high-level video intents into executable operations.



\begin{table*}[t]
\centering
\caption{
Representative video-agent tool libraries compared by tool granularity, functionality, and organization.
\toolyes and \toolno denote supported and unsupported features, respectively.
N/R denotes that the number of tools is not reported.
}
\label{tab:tool_library_comparison}
\footnotesize
\resizebox{\textwidth}{!}{%
\begin{tabular}{lcccccc}
\toprule
\textbf{Work} & \textbf{\# Tools} & \textbf{Tool Type} & \textbf{Audio} & \textbf{Function-aware Routing} & \textbf{Fine-grained Meta Tools} & \textbf{Registry Support} \\
\midrule
ProViQ~\citep{choudhury2024video} & N/R & Program APIs & \toolno & \toolno & \toolno & \toolno \\
DoraemonGPT~\citep{yang2024doraemongpt} & N/R & Subtask / Knowledge & \toolyes & \toolno & \toolno & \toolno \\
VideoAgent~\citep{fan2024videoagent} & 4 & Caption / Locate / VQA / Object & \toolno & \toolno & \toolno & \toolno \\
DVD~\citep{zhang2025deep} & 3 & Global Browse / Clip Search / Frame Inspect & \toolyes & \toolno & \toolno & \toolno \\
VITAL~\citep{zhang2025thinking} & 3 & Clip Caption / Clip QA / Video Clip & \toolno & \toolno & \toolno & \toolno \\
Ego-R1~\citep{tian2025ego} & 3 & H-RAG / Video-LLM / VLM & \toolyes & \toolno & \toolno & \toolno \\
STAR~\citep{fan2025tool} & 22 & Spatial / Temporal / General & \toolno & \toolyes & \toolno & \toolno \\
VideoThinker~\citep{li2026videothinker} & 6 & Temporal Retrieval / Temporal Zoom & \toolyes & \toolyes & \toolno & \toolno \\
VideoSeek~\citep{lin2026videoseek} & 3 & Overview / Skim / Focus & \toolno & \toolyes & \toolno & \toolno \\
EGAgent~\citep{rege2026agentic} & 4 & Visual / Audio Transcript / Entity Graph / Analyzer & \toolyes & \toolyes & \toolno & \toolno \\
\rowcolor{retoolrow}
\midrule
\textsc{ReTool-Video} (Ours) & 134 & Base / Fine-grained meta tools & \toolyes & \toolyes & \toolyes & \toolyes \\
\bottomrule
\end{tabular}%
}
\vspace{-5mm}
\end{table*}

\begin{wrapfigure}[16]{r}{0.56\textwidth}
    \centering
    \includegraphics[width=\linewidth]{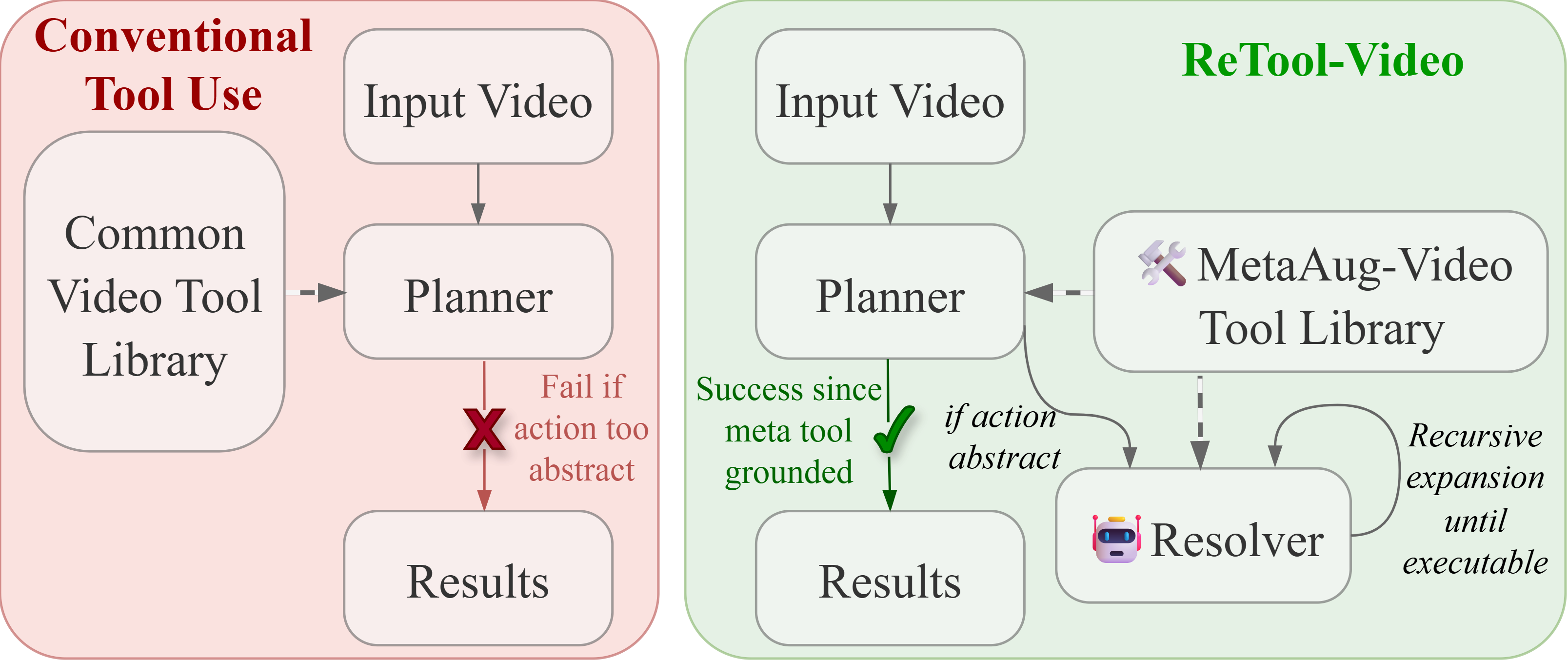}
    \caption{Motivation of \textsc{ReTool-Video}. Conventional video tool-using agents force abstract actions (e.g., temporal merging or cross-modal verification) into primitive calls, while \textsc{ReTool-Video} recursively grounds them into executable multimodal tool chains.}
    \label{fig:teaser}
\vspace{-0.3cm}
\end{wrapfigure}



Despite these advances, existing video-based agents still face two core challenges in scaling tool use for complex video understanding. First, 
\textbf{from the agent tool-level view, existing video agents lack a broad and composable tool space.}
Video reasoning requires not only tools for acquiring evidence from frames, clips, audio, subtitles, or scene graphs, but also meta operations for filtering, temporal merging, counting, reranking, computation, and formatting. These operations are necessary to decompose high-level video intents into executable steps. 
However, as shown in Table~\ref{tab:tool_library_comparison}, existing video agents mostly rely on a few coarse tools for retrieval, localization, VQA, or clip inspection~\citep{choudhury2024video,zhang2025deep,lin2026videoseek}, with limited meta-level support such as filtering, temporal merging, aggregation, and format conversion.~\citep{rege2026agentic}.


Secondly, \textbf{from the agent action-level view, existing video agents force high-level video intents into primitive tool calls.} As shown in Figure~\ref{fig:teaser}, most tool-use methods require each action to be directly mapped to a registered executable tool with valid parameters~\citep{yao2022react,schick2023toolformer}, and video agents inherit this flat interface for retrieval, localization, frame inspection, or clip QA~\citep{fan2024videoagent,zhang2025thinking,lin2026videoseek}. Yet many video intents, such as checking whether adjacent clips form one continuous event, comparing object states across time, or aggregating repeated actions, are abstract and compositional. When they do not match a single registered tool, flat tool-calling often leads to ill-suited tool choices, brittle parameters, or premature termination, motivating a runtime mechanism that grounds high-level video intents into executable multimodal operations.

To address the tool-level challenge, we construct a MetaAug-Video Tool Library (MVTL) that expands the capability space of video agents. MVTL is an independent, retrievable, and composable library with 134 registered tools, including 26 base tools for general multimodal signal processing and 108 meta tools for filtering, aggregation, reranking, computation, formatting, and other intermediate-result operations. It supports dual-level access to both structured video information, such as subtitles, captions, event descriptions, and knowledge graphs, and raw modal evidence, such as frames and video clips, providing a hierarchical and multifunctional tool space for diverse video scenarios. Moreover, MVTL is extensible by design: new domain-specific or task-specific tools can be plugged into the library through the same registration schema, enabling the tool space to be expanded without modifying the main reasoning framework.
In addition, to address the action-level challenge, we propose \textsc{ReTool-Video}, a recursive tool-using method that grounds high-level video intents into executable operations. Given MVTL as the tool environment, matched actions are executed directly, while unmatched actions are treated as abstract video-reasoning intents and delegated to a resolver, which grounds them into executable tool chains through parameter repair, tool substitution, or decomposition. We further train the planner with reinforcement learning in the same execution loop used at inference time. Experiments on MVBench, MLVU, and Video-MME$_{\mathrm{w/o\ sub.}}$ show that \textsc{ReTool-Video} consistently outperforms other strong baselines, and further analysis demonstrates that meta tools provide additional gains by enabling fine-grained filtering, aggregation, temporal merging, and post-processing over intermediate results. The contributions of this paper are summarized as follows:

\begin{itemize}[leftmargin=*]
\item We construct MVTL, a video-agent tool library with 134 registered tools, covering base tools for multimodal evidence acquisition and meta tools for fine-grained intermediate-result processing.

\item We propose \textsc{ReTool-Video}, a recursive tool-using method that resolves high-level video intents by directly executing matched actions and grounding unmatched intents into executable tool chains.

\item We evaluate \textsc{ReTool-Video} on three widely used benchmarks, demonstrating strong performance over competitive baselines and validating the benefits of recursive grounding and meta tools.

\end{itemize}

\section{Related Work}
\subsection{Video Understanding and Video Reasoning}
Video understanding has evolved from short-clip action recognition to tasks requiring temporal localization, compositional QA, causal-temporal reasoning, state tracking, action counting, and cross-modal evidence alignment~\citep{krishna2017dense}. Benchmarks such as TVQA~\citep{lei2018tvqa} and NExT-QA~\citep{xiao2021next} show that many questions require localizing moments, understanding dialogue or subtitles, comparing actions over time, and grounding answers in temporal evidence. Recent video-language models improve video QA through alignment, instruction tuning, and unified image-video representations~\citep{lin2024video,maaz2024video}, while long-video methods expand temporal coverage with sparse memory, long-context transfer, document-style retrieval, or hierarchical indexing~\citep{song2024moviechat,zhang2024long,ma2025drvideo,yin2025videoarm}. Evidence-oriented systems further retrieve candidate windows and verify them with clip-, frame-, audio-, transcript-, or event-level observations~\citep{zhang2025deep,zhang2025thinking}. However, most methods focus on evidence access or context compression, with limited support for meta-level operations such as filtering, temporal merging, counting, reranking, computation, and formatting. To this end, we introduce MetaAug-Video, which combines base tools for multimodal evidence acquisition with meta tools for intermediate-result processing, and \textsc{ReTool-Video}, which grounds high-level video intents into executable operations.

\subsection{Agentic Reasoning and Tool-Augmented Multimodal Systems}

Agentic reasoning extends large models from direct response generation to multi-step interaction, where the model plans actions, invokes tools, observes feedback, and updates subsequent decisions~\citep{yao2022react}. General tool-use agents show that external APIs, expert models, and task-specific tools can expand model capabilities beyond parametric knowledge~\citep{schick2023toolformer,shen2023hugginggpt}. Multimodal agents further connect language models with visual experts, OCR, detection, segmentation, executable code, or visual programs for compositional reasoning over perception modules and intermediate results~\citep{yang2023mm,suris2023vipergpt}. For video tasks, recent agents introduce planners, retrievers, memory modules, temporal grounding, frame or clip inspection, verifiers, and tool-guided seeking for long-horizon evidence search and multi-step verification~\citep{fan2024videoagent,liu2025videomind,zhang2025deep,zhang2025thinking,lin2026videoseek}.
However, many video agents still rely on a flat tool-calling interface, where each action must be mapped to a registered executable tool and the observation is folded back into the next step~\citep{fan2024videoagent,zhang2025deep,lin2026videoseek}. This is restrictive because video reasoning often involves abstract and compositional intents, such as merging adjacent detections, comparing states across time, or aggregating repeated events, which may not correspond to a single tool. Although prior multi-agent and recursive planning studies explore decomposition, role separation, and parallel scheduling~\citep{li2024agent,liu2025pc,yu2025recode,wu2025gap}, \textsc{ReTool-Video} targets video-specific intent grounding: it directly executes matched actions and delegates unmatched intents to a resolver that grounds them into executable tool chains over \textsc{MetaAug-Video}.

\section{\textsc{MVTL}: MetaAug-Video Tool Library}
\label{sec:tool_library}

\begin{figure}[t!]
    \centering
    \begin{minipage}[c]{0.38\linewidth}
        \centering
        \footnotesize
        \setlength{\tabcolsep}{5pt}
        \renewcommand{\arraystretch}{1.5}
        \begin{tabular}{lc}
        \toprule
        \textbf{Base Tool Function} & \textbf{\#} \\
        \midrule
        Retrieval/Search & 10 \\
        Visual/Video & 7 \\
        Audio/Speech & 4 \\
        Execution/Coding & 3 \\
        Memory/System & 2 \\
        \bottomrule
        \end{tabular}
    \end{minipage}
    \hfill
    \begin{minipage}[c]{0.58\linewidth}
        \centering
        \includegraphics[width=\linewidth]{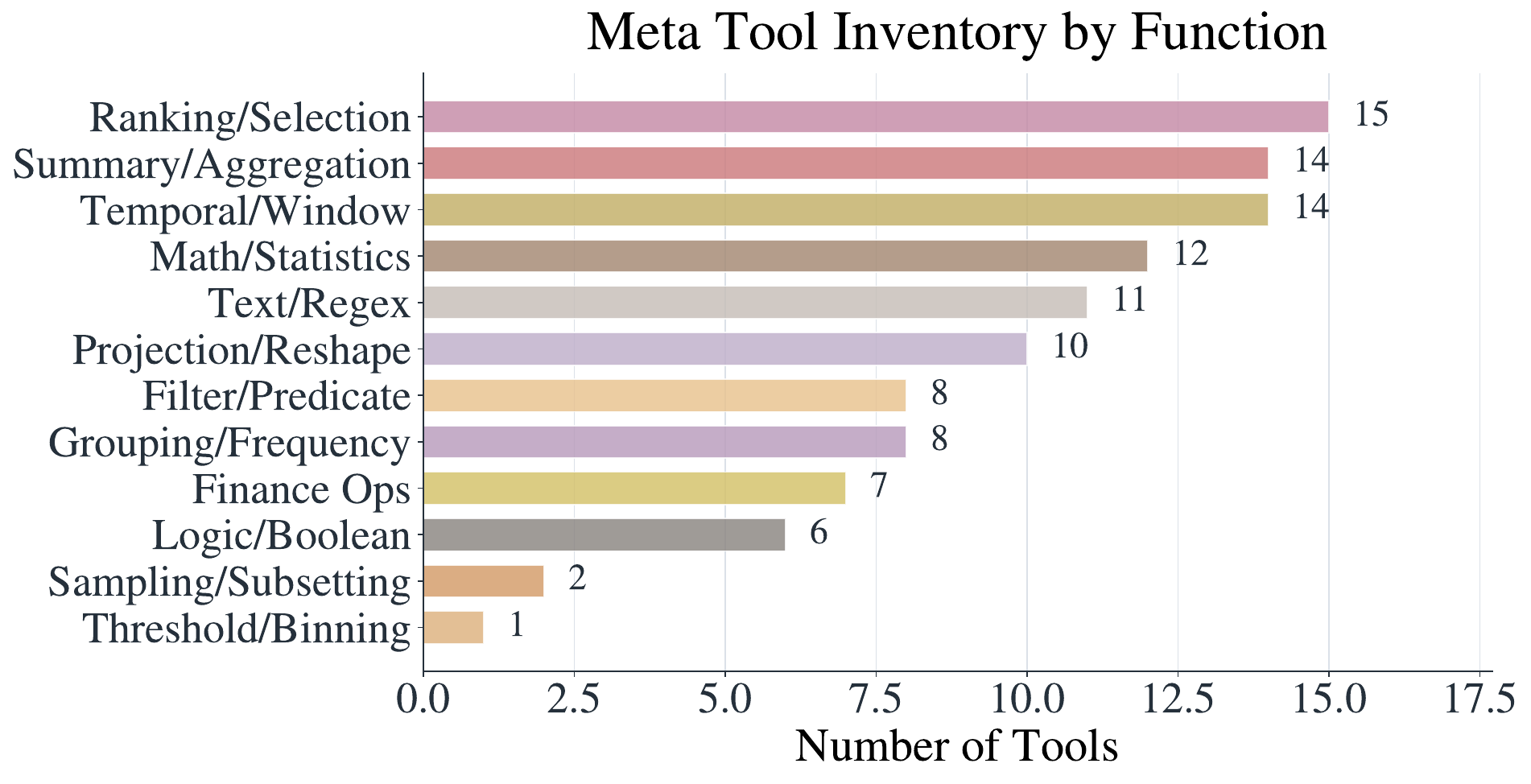}
    \end{minipage}
    \vspace{-3mm}
    \caption{
    Functional categories of base tools and meta tools in \textsc{MVTL}. 
    }
    \label{fig:toolset}
    \vspace{-7mm}
\end{figure}

Complex video reasoning requires more than coarse interfaces for retrieval, localization, or clip-level QA. Beyond acquiring multimodal evidence, an agent also needs to organize intermediate results through operations such as filtering, temporal merging, counting, reranking, computation, and formatting. As shown in Table~\ref{tab:tool_library_comparison}, existing video-agent tool libraries usually provide limited functional coverage and lack a structured tool space for compositional execution. To address this tool-level challenge, we construct \textsc{MVTL}, a MetaAug-Video Tool Library that combines base tools for multimodal information access with meta tools for intermediate-result processing. \textsc{MVTL} is an independent, retrievable, and extensible library that serves as the capability foundation for recursive video-agent reasoning. 

\textbf{Base and Meta Tools.}
\textsc{MVTL} contains 134 registered tools, including 26 base tools and 108 meta tools. Base tools access videos, images, audio, transcripts, scene graphs, and structured resources, supporting temporal retrieval, frame or clip inspection, audio/ASR access, graph exploration, visual analysis, and cross-modal verification. Meta tools operate on outputs from base tools or previous reasoning steps, providing atomic operations for filtering, temporal merging, counting, aggregation, reranking, text processing, format conversion, and lightweight computation. This base--meta design prevents complex video intents from being collapsed into a single coarse tool call, allowing them to be grounded into executable chains that acquire, refine, and compose multimodal information.

\textbf{Functional Coverage.}
Beyond the base--meta distinction, \textsc{MVTL} organizes tools by function, as shown in Figure~\ref{fig:toolset}. Base tools cover major video-agent capabilities, including retrieval/search, visual/video analysis, audio/speech processing, execution, memory/system operations, and domain-specific extensions. Meta tools cover common intermediate-result operations, including ranking, aggregation, temporal/window processing, mathematical and text operations, filtering, grouping, sampling, and thresholding. The implementations include both programmatic tools for deterministic operations and model-based tools for bounded perception or semantic transformation, with details in Appendix~\ref{app:model_based_tools}. This functional organization makes the tool space broad but navigable, supporting executable chains that combine base and meta operations.

\textbf{Dual-Level Video Access.}
\textsc{MVTL} supports two complementary levels of video access. The structured level provides preprocessed information such as subtitles, captions, event descriptions, and knowledge graphs for efficient localization, retrieval, and candidate-window generation. The raw-evidence level provides timestamped frames and video clips for more specialized, fine-grained, and information-rich visual evidence acquisition. This design balances efficiency and fidelity: the agent can first use structured signals to narrow the search space and then use raw modal evidence for faithful verification.

\textbf{Registry and Routing.}
\textsc{MVTL} is implemented as a registry-based capability layer rather than a loose script collection. Each tool is registered with a description, tags, input schema, output format, availability conditions, and runtime constraints. During execution, the planner may specify a tool name, capability description, or partially specified intent. The routing layer retrieves candidate tools, validates schemas and availability, executes matched tools, and normalizes outputs into a shared observation format. When parameters are incomplete or outputs are invalid, the runtime applies bounded repair, retry, or fallback. If no executable tool is matched, the action is passed to the \textsc{ReTool-Video} resolver for recursive grounding. This separates capability organization from safe execution, allowing \textsc{MVTL} to expand while keeping the planner interface stable.

\begin{figure}[t!]
    \centering
    \includegraphics[width=1\linewidth]{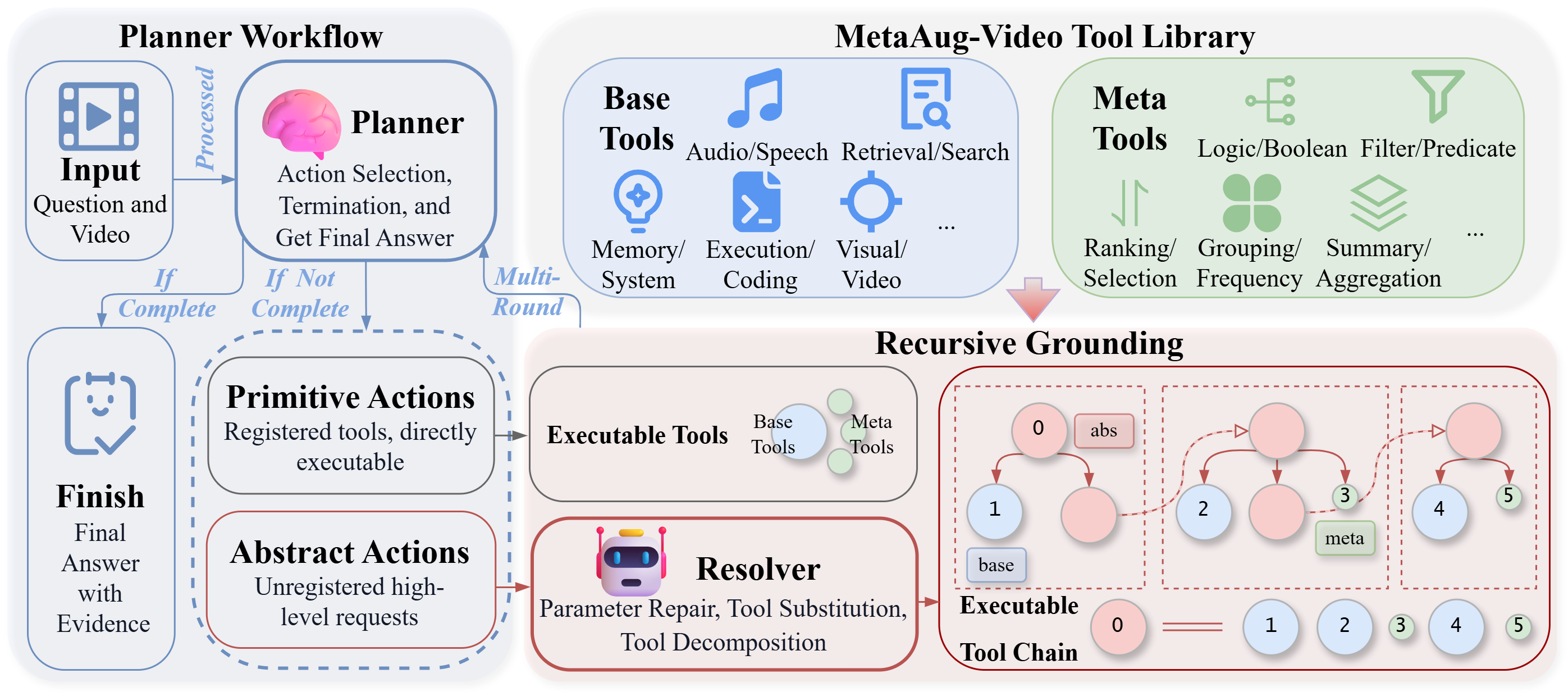}
\vspace{-6mm}
    \caption{Overall framework of \textsc{ReTool-Video}, where the planner selects primitive or abstract actions, the MetaAug-Video tool library provides base and meta tools, and the resolver recursively grounds abstract actions into executable tool chains.}
    \label{fig:loop}
\vspace{-5mm}
\end{figure}

\section{\textsc{ReTool-Video}}

\subsection{Recursive Tool Grounding}

Complex video QA can be viewed as an interactive decision process, where an agent repeatedly determines what information is missing, invokes tools to acquire or process it, and decides whether the accumulated observations are sufficient for answering. The overall framework is shown in Figure~\ref{fig:loop}. We define each task instance as
\begin{equation}
x=(\mathcal{V}, q, c_0),
\end{equation}
where $\mathcal{V}$ denotes one or more videos, $q$ is the question, and $c_0$ is the initial context. The system produces an interaction trajectory
\begin{equation}
\tau = \{(s_t,a_t,o_t)\}_{t=1}^{T},
\end{equation}
where $s_t$ contains the task objective, accumulated observations, execution history, and budget state; $a_t$ is the planner action; and $o_t$ is the observation returned by tools or the environment, including textual outputs, timestamps, visual evidence, and execution signals.

The key design of \textsc{ReTool-Video} is to separate \textit{intent expression} from \textit{immediate tool executability}. Instead of requiring every planner action to match a registered tool exactly, we allow the planner to express high-level video-reasoning intents. The runtime then either executes them directly when they match the tool registry, or recursively grounds them into executable tool chains when they are abstract or underspecified.

\begin{wrapfigure}[19]{r}{0.48\textwidth}
\vspace{-0.9cm}
    \centering
    \includegraphics[width=0.95\linewidth]{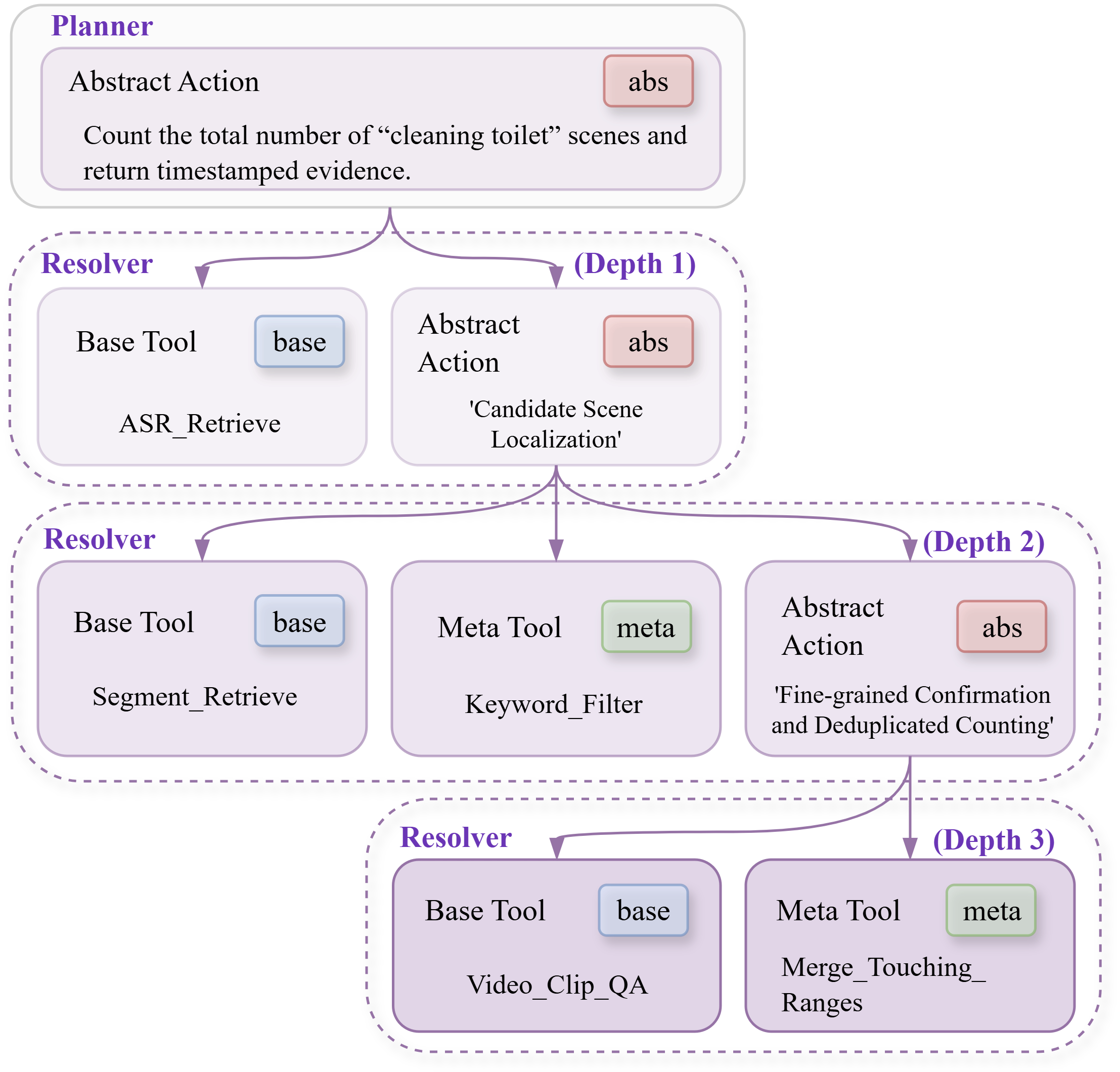}
    \caption{\small An example of recursive tool grounding in \textsc{ReTool-Video}, where abstract actions are recursively resolved into executable tool chains.}
    \label{fig:recursive}
\vspace{0.2cm}
\end{wrapfigure}

\textbf{Action Interface.}
At each step, the planner outputs either a set of structured actions or a top-level \textsc{Finish}. The action space contains three types:
\begin{equation}
\label{eq:action_space}
\mathcal{A} = \mathcal{A}_{\mathrm{prim}} \cup \mathcal{A}_{\mathrm{abs}} \cup Finish,
\end{equation}
where $\mathcal{A}_{\mathrm{prim}}$ denotes primitive actions that can be executed by registered tools, $\mathcal{A}_{\mathrm{abs}}$ denotes abstract video-reasoning intents that require runtime grounding, and \textit{Finish} denotes task termination with the final answer. This action interface is useful for video reasoning because many operations, such as comparing states across time, merging adjacent events, or aggregating repeated actions, are natural reasoning intents but may not correspond to a single tool in the registry.

\textbf{Direct Execution and Recursive Grounding.}
Given a planner action $a_t$, the runtime first queries the tool registry in \textsc{MVTL}. If $a_t$ matches a registered tool and satisfies its schema constraints, it is executed directly:
\begin{equation}
o_t = \mathrm{Exec}(a_t), \quad a_t \in \mathcal{A}_{\mathrm{prim}}.
\end{equation}
If no executable tool can be matched, \textsc{ReTool-Video} treats the action as an abstract intent rather than an execution failure. The intent is delegated to a resolver, which grounds it through parameter repair, tool substitution, or decomposition into lower-level tool calls:
\begin{equation}
o_t = \mathrm{Resolver}(a_t), \quad a_t \in \mathcal{A}_{\mathrm{abs}}.
\end{equation}
For example, an intent such as checking whether adjacent clips form the same event can be resolved into clip inspection, visual comparison, temporal merging, and result aggregation. The resolver returns its local results as structured observations to the root planner, but it is not allowed to output the final answer. Only the root planner can emit \textsc{Finish}, which separates local tool grounding from global evidence-sufficiency judgment.

\textbf{Controlled Parallel Execution.}
The planner may issue multiple independent actions in one round. \textsc{ReTool-Video} executes these actions under controlled parallelism when their dependencies allow it. This enables the agent to probe multiple temporal windows, inspect different modalities, or retrieve evidence from different entity clues without forcing all operations into a strictly serial chain. After execution, all tool outputs are normalized into a shared observation format and aggregated at the round boundary before being passed to the next planner state. Together, direct execution, recursive grounding, and controlled parallelism allow \textsc{ReTool-Video} to resolve high-level video intents into executable multimodal operations while preserving a stable global reasoning loop.

\subsection{Reinforcement Learning on Planner}
\label{sec:planner_rl}

We optimize the planner policy with reinforcement learning, while treating the resolver, tool library, and other execution modules as non-trainable components. This focuses the learning signal on high-level decisions, including action selection, abstract-intent delegation, evidence sufficiency judgment, and termination. Given a task instance \(x=(\mathcal{V},q,c_0)\), the old planner policy
\(\pi_{\mathrm{old}}\) samples a group of \(G\) complete episodes
\(\{\tau_i\}_{i=1}^{G}\). Each episode is scored by a trajectory-level reward, and the rewards within
the same group are used to compute relative advantages for planner-policy optimization.

\textbf{Reward Design.} We use final answer correctness as the dominant reward, with only lightweight
auxiliary terms for structural validity and ineffective trajectories:
\begin{equation}
R(\tau)
=
R_{\mathrm{ans}}(\tau)
+
\lambda_{\mathrm{valid}} C_{\mathrm{valid}}(\tau)
-
\lambda_{\mathrm{cost}} C_{\mathrm{cost}}(\tau).
\end{equation}
For multiple-choice questions, \(R_{\mathrm{ans}}\) is binary; for open-ended questions, it follows the
dataset-specific evaluator and is normalized to \([0,1]\):
\begin{equation}
R_{\mathrm{ans}}(\tau)
=
\begin{cases}
\mathbf{1}[\hat{y}=y], & \text{for multiple-choice QA},\\
\mathrm{Eval}_{\mathcal{D}}(\hat{y},y), & \text{for open-ended QA}.
\end{cases}
\end{equation}
Here, \(C_{\mathrm{valid}}\) indicates whether the episode satisfies the required JSON format, action
syntax, and finish protocol, while \(C_{\mathrm{cost}}\) penalizes clearly ineffective trajectories such
as empty progress, repeated probing, or max-step termination.

\textbf{Group-relative Computation.} For each sampled episode \(\tau_i\), let \(R_i=R(\tau_i)\). The
episode-level advantage is computed by normalizing rewards within the same group:
\begin{equation}
\hat{A}_i
=
\frac{
R_i-\mu(\{R_j\}_{j=1}^{G})
}{
\sigma(\{R_j\}_{j=1}^{G})+\epsilon
}.
\end{equation}
This corresponds to the RL branch in Figure~\ref{fig:loop}: sampled episodes are first
evaluated by trajectory rewards, then compared through group computation, and finally used to
update the planner.

\textbf{Planner-policy Objective.} Let \(\mathcal{T}^{p}_{i}\) denote the planner-token positions in episode
\(\tau_i\), and let \(\rho_{i,t}(\theta)\) be the probability ratio between the updated planner policy
\(\pi_{\theta}\) and the old policy \(\pi_{\mathrm{old}}\). We optimize the planner with a GRPO~\citep{shao2024deepseekmath} clipped objective:
\begin{equation}
\max_{\theta} J(\theta)
=
\mathbb{E}_{x\sim\mathcal{D},\,\{\tau_i\}_{i=1}^{G}\sim\pi_{\mathrm{old}}}
\left[
\frac{1}{G}
\sum_{i=1}^{G}
\sum_{t\in\mathcal{T}^{p}_{i}}
L^{\mathrm{clip}}_{i,t}(\theta)
\right]
-
\beta D_{\mathrm{KL}}^{\mathrm{rev}}
\left(
\pi_{\theta}\Vert \pi_{\mathrm{ref}}
\right),
\end{equation}
where \(\pi_{\mathrm{ref}}\) is a frozen reference VLM used for KL regularization. The clipped
token-level objective is
\begin{equation}
L^{\mathrm{clip}}_{i,t}(\theta)
=
\min
\left(
\rho_{i,t}(\theta)\hat{A}_i,\,
\mathrm{clip}
\left(
\rho_{i,t}(\theta),
1-\epsilon_{\mathrm{low}},
1+\epsilon_{\mathrm{high}}
\right)
\hat{A}_i
\right).
\end{equation}
Only planner tokens are included in the policy loss, while resolver tokens, tool outputs,
observations, and environment-generated text are masked out. Therefore, RL updates the planner's
global evidence-orchestration policy rather than the local resolver or tool execution process.
Additional implementation details are provided in Appendix~\ref{app:rl_details}.

\begin{table}[!htbp]
\centering
\caption{Main results comparing \textsc{ReTool-Video} with representative large video language models.}
\label{tab:main_results}
\scriptsize
\setlength{\tabcolsep}{3.5pt}
\renewcommand{\arraystretch}{0.92}
\begin{tabular}{l|c|ccc}
\toprule
\textbf{Models} & \textbf{Size} & \textbf{MLVU} & \textbf{MVBench} & \textbf{Video-MME$_{\mathrm{w/o\ sub.}}$} \\
\midrule
\multicolumn{5}{l}{\textit{Closed-source models}} \\
\midrule
GPT4-V~\citep{openai2023gpt4v} & - & - & 43.5 & 60.7 \\
GPT-5-Nano-2025-08-07~\citep{singh2025openai} & - & 69.2 & - & 66.2 \\
GPT-4o~\citep{hurst2024gpt} & - & 64.6 & - & 71.9 \\
Claude-Sonnet-4.5~\citep{anthropic2025claudesonnet45systemcard} & - & 72.8 & - & 75.3 \\
Gemini-2.5-Flash-Lite~\citep{comanici2025gemini} & - & 78.5 & - & 72.7 \\
\midrule
\multicolumn{5}{l}{\textit{Open-source models}} \\
\midrule
LLaMA-VID~\citep{li2024llama} & 7B & 33.2 & 41.9 & - \\
ShareGPT4Video~\citep{chen2024sharegpt4video} & 8B & 33.8 & 51.2 & 43.6 \\
VideoLLaMA2~\citep{cheng2024videollama} & 8$\times$7B & 45.6 & 54.6 & 46.6 \\
VideoThinker~\citep{li2026videothinker} & 7B & 54.8 & - & 53.7 \\
Framemind~\citep{ge2025framemind} & 7B & 48.6 & 64.2 & 60.9 \\
VideoExplorer~\citep{yuan2025videoexplorer} & 7B\&32B & 58.6 & - & - \\
Kangaroo~\citep{liu2026kangaroo} & 8B & 61.0 & 61.0 & 56.0 \\
VideoMind~\citep{liu2025videomind} & 7B & 64.4 & - & 58.2 \\
Flash-VStream~\citep{zhang2025flash} & 7B & 66.3 & 65.4 & 61.2 \\
DVD~\citep{zhang2025deep} & - & - & - & 67.3 \\
NVILA~\citep{liu2025nvila} & 8B & 70.1 & 68.1 & 64.2 \\
STAR~\citep{fan2025tool} & - & - & - & 70.0 \\
VideoSeek~\citep{lin2026videoseek} & - & - & - & 70.1 \\
InternVL3.5-30B-A3B~\citep{wang2025internvl3_5} & 30B & 73.0 & 72.1 & 68.7 \\
\rowcolor{retoolrow}
\textsc{ReTool-Video} (Ours) & 9B & \textbf{81.5} & \textbf{72.9} & \textbf{76.6} \\
\bottomrule
\end{tabular}
\vspace{-5mm}
\end{table}

\section{Experiments}

\subsection{Experiment Settings}
\textbf{Benchmarks and Evaluation Metrics.}
We evaluate \textsc{ReTool-Video} on three representative general-purpose video understanding benchmarks: MVBench~\citep{li2024mvbench}, MLVU~\citep{zhou2025mlvu}, and Video-MME$_{\mathrm{w/o\ sub.}}$~\citep{fu2025video}.  MVBench focuses on short-video temporal understanding, while MLVU emphasizes long-video reasoning with tasks covering both global video comprehension and local evidence reasoning. Video-MME provides a comprehensive open-domain evaluation across diverse video durations and scenarios; we use its standard no-subtitle setting, denoted as Video-MME$_{\mathrm{w/o\ sub.}}$. In the experiment, we follow the official evaluation protocols and report \textit{accuracy} as the main metric. More details about benchmarks and evaluations are listed in Appendix~\ref{app:benchmark_descriptions}.


\textbf{Baselines and Implementation Details.}
We compare \textsc{ReTool-Video} with closed-source and open-source baselines. Closed-source models include GPT4-V~\citep{openai2023gpt4v}, GPT-5-Nano~\citep{singh2025openai}, GPT-4o~\citep{hurst2024gpt}, Claude-Sonnet-4.5~\citep{anthropic2025claudesonnet45systemcard}, and Gemini-2.5-Flash-Lite~\citep{comanici2025gemini}. Open-source baselines cover general video-language models, including LLaMA-VID~\citep{li2024llama}, ShareGPT4Video~\citep{chen2024sharegpt4video}, VideoLLaMA2~\citep{cheng2024videollama}, Kangaroo~\citep{liu2026kangaroo}, Flash-VStream~\citep{zhang2025flash}, NVILA~\citep{liu2025nvila}, and InternVL3.5~\citep{wang2025internvl3_5}, as well as video reasoning or agentic methods, including Framemind~\citep{ge2025framemind}, VideoMind~\citep{liu2025videomind}, VideoExplorer~\citep{yuan2025videoexplorer}, VideoThinker~\citep{li2026videothinker}, DVD~\citep{zhang2025deep}, STAR~\citep{fan2025tool}, and VideoSeek~\citep{lin2026videoseek}.
In the experiment, \textsc{ReTool-Video} uses Qwen/Qwen3.5-9B~\citep{qwen3.5} as both planner and resolver, and all baselines utilize their official processing method. Detailed settings for our framework and all baselines, including frame sampling, retrieval/tool budgets, subtitle usage, and preprocessing protocols, are provided in Appendix~\ref{app:baseline_protocols}.

\vspace{-5pt}
\subsection{Main Results}
Table~\ref{tab:main_results} reports the main results of \textsc{ReTool-Video} against both closed-source and open-source video-language models. \textsc{ReTool-Video} achieves 72.9 on MVBench, 81.5 on MLVU, and 76.6 on Video-MME$_{\mathrm{w/o\ sub.}}$, showing strong performance across short-video, long-video, and subtitle-free video understanding settings. Compared with InternVL3.5-30B-A3B, \textsc{ReTool-Video} improves by 8.5\% on MLVU and 7.9\% on Video-MME$_{\mathrm{w/o\ sub.}}$ despite using a smaller backbone. It also outperforms strong open-source baselines such as NVILA, Flash-VStream, and VideoMind, with particularly large gains on MLVU, where questions often require evidence retrieval and integration over extended temporal horizons. The improvement on MVBench is more moderate, likely because many MVBench questions involve shorter videos and local temporal perception, where strong VLMs can already answer from sampled visual context. In contrast, the larger gains on MLVU and Video-MME$_{\mathrm{w/o\ sub.}}$ suggest that \textsc{ReTool-Video} is most effective when answers depend on sparse evidence, long-range temporal search, cross-segment aggregation, or multimodal verification. These results indicate that the bottleneck in complex video QA is not solely model scale or context length; reliable tool use and adaptive evidence orchestration are also critical for strong video reasoning.

\begin{table}[!t]
\centering
\caption{Ablation study results of \textsc{ReTool-Video}.}
\label{tab:ablation}
\scriptsize
\setlength{\tabcolsep}{4pt}
\renewcommand{\arraystretch}{0.9}
\begin{tabular}{lccc}
\toprule
\textbf{Setting} & \textbf{MLVU} & \textbf{MVBench} & \textbf{Video-MME$_{\mathrm{w/o\ sub.}}$} \\
\midrule
Direct Response (w/o Tools) & 47.6 & 46.6 & 57.4 \\
\midrule
Tool-use w/o Parallel\&Recursion & 70.5 & 63.5 & 71.8 \\
Tool-use w/o Meta Tools & 71.2 & 64.1 & 71.7 \\
Tool-use w/o Parallel & 68.4 & 64.4 & 70.2 \\
Tool-use w/o Recursion & 63.2 & 64.8 & 66.7 \\
Tool-use & 74.6 & 65.3 & 73.3 \\
\midrule
\textsc{ReTool-Video} & \textbf{81.5} & \textbf{72.9} & \textbf{76.6} \\
\bottomrule
\end{tabular}
\vspace{-3mm}
\end{table}

\subsection{Ablation Study}

\begin{wrapfigure}[16]{r}{0.50\textwidth}
\vspace{-1.7cm}
    \centering
    \includegraphics[width=\linewidth]{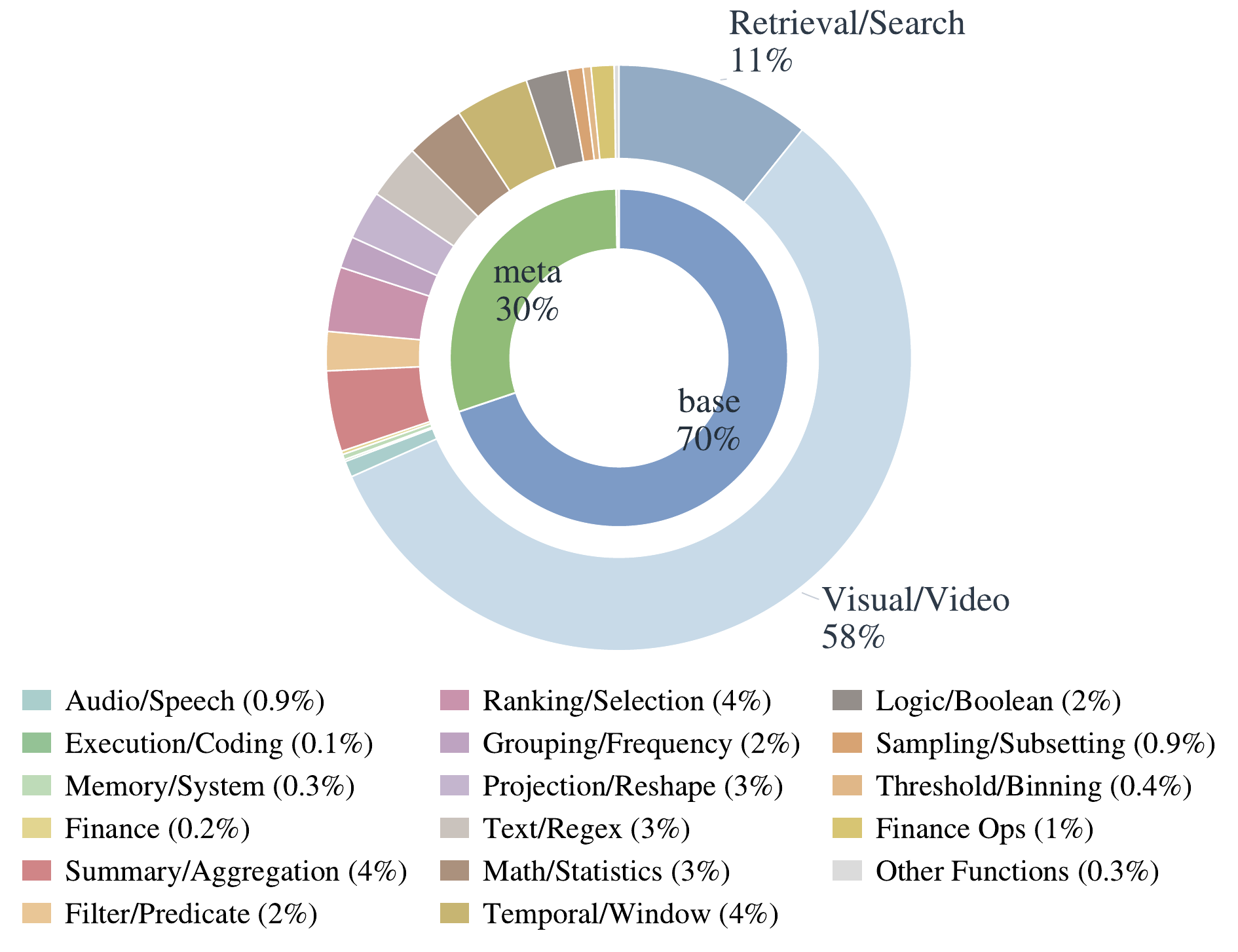}
\vspace{-4mm}
    \caption{Distribution of runtime tool calls in \textsc{ReTool-Video}, showing the relative invocation frequency of each base/meta tool category during model inference.}
    \label{fig:tool_usage}
\vspace{-0.2cm}
\end{wrapfigure}

To assess each component, we conduct the ablation study. As shown in Table~\ref{tab:ablation}, {Direct Response} is a no-tool baseline, all ``w/o'' variants are ablated from the full {Tool-use} framework without planner-level RL, and \textsc{ReTool-Video} denotes the full system with RL. From the results, we could find: (1) active tool use substantially improves over direct answering, especially on long-video benchmarks such as MLVU, confirming the importance of retrieval, local verification, and evidence aggregation for complex video QA. (2) meta tools bring consistent gains over the variant without meta-level operations, showing the value of filtering, aggregation, reranking, and other intermediate-result processing beyond general multimodal signal processing. (3) recursive grounding and parallel execution also provide structural benefits: removing recursion while retaining parallelism can underperform removing both on MLVU and Video-MME$_{\mathrm{w/o\ sub.}}$, suggesting that parallel calls are effective only when abstract actions can be reliably grounded; otherwise, they may introduce noisy observations and context pressure. (4) planner-level RL further improves over {Tool-use}, indicating that action selection, evidence sufficiency estimation, and termination remain key bottlenecks.

\begin{figure}[t!]
    \centering
    \includegraphics[width=1\linewidth]{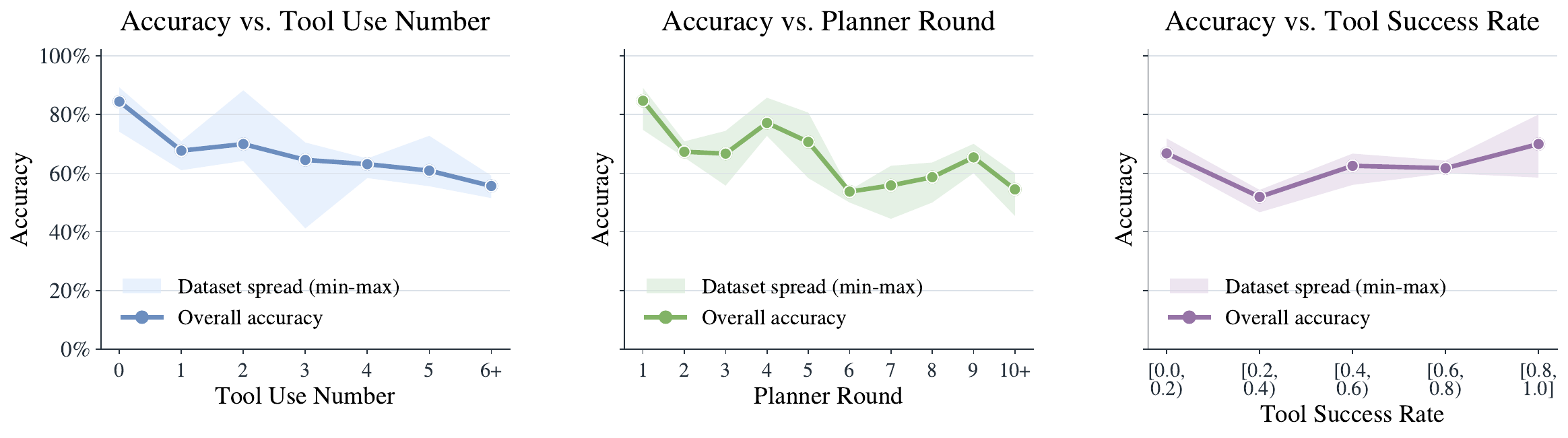}
\vspace{-7mm}
    \caption{Relationship between final accuracy and tool-use behavior, including tool-call count, reasoning iterations, and tool-call success rate.}
    \label{fig:curves}
\vspace{-5mm}
\end{figure}

\subsection{Tool-Behavior Analysis}

To better understand the tool utilization behavior of \textsc{ReTool-Video}, we analyze final accuracy against three trajectory-level statistics: tool-call count, reasoning iterations, and tool-call success rate. As shown in Figure~\ref{fig:curves}, the solid curve reports average accuracy over three benchmarks, and the shaded region denotes the min--max range. The results suggest that effective tool use depends more on quality than quantity. Accuracy remains stable with a small to moderate number of tool calls, especially one to four calls, but does not improve monotonically as tool chains become longer. A similar trend appears for reasoning iterations: moderate multi-turn reasoning helps localize, verify, and aggregate evidence, while overly long trajectories often indicate unresolved uncertainty or repeated probing. Tool-call success rate is more strongly correlated with accuracy than raw tool frequency, indicating that usable observations are more important than simply invoking more tools.

Figure~\ref{fig:tool_usage} further shows the runtime tool distribution. Base tools dominate, reflecting the central role of first-hand multimodal evidence acquisition, while meta tools still account for a meaningful share of executions for filtering, aggregation, temporal merging, memory management, and lightweight computation. By function, visual/video tools are most frequently used, followed by retrieval/search and auxiliary groups. This pattern matches the design of \textsc{MetaAug-Video}, where high-frequency base tools handle common evidence acquisition, while meta and auxiliary tools support long-tail reasoning and intermediate-result processing over accumulated observations.

\subsection{Case Study}
\begin{figure}[t!]
    \centering
    \includegraphics[width=0.95\linewidth]{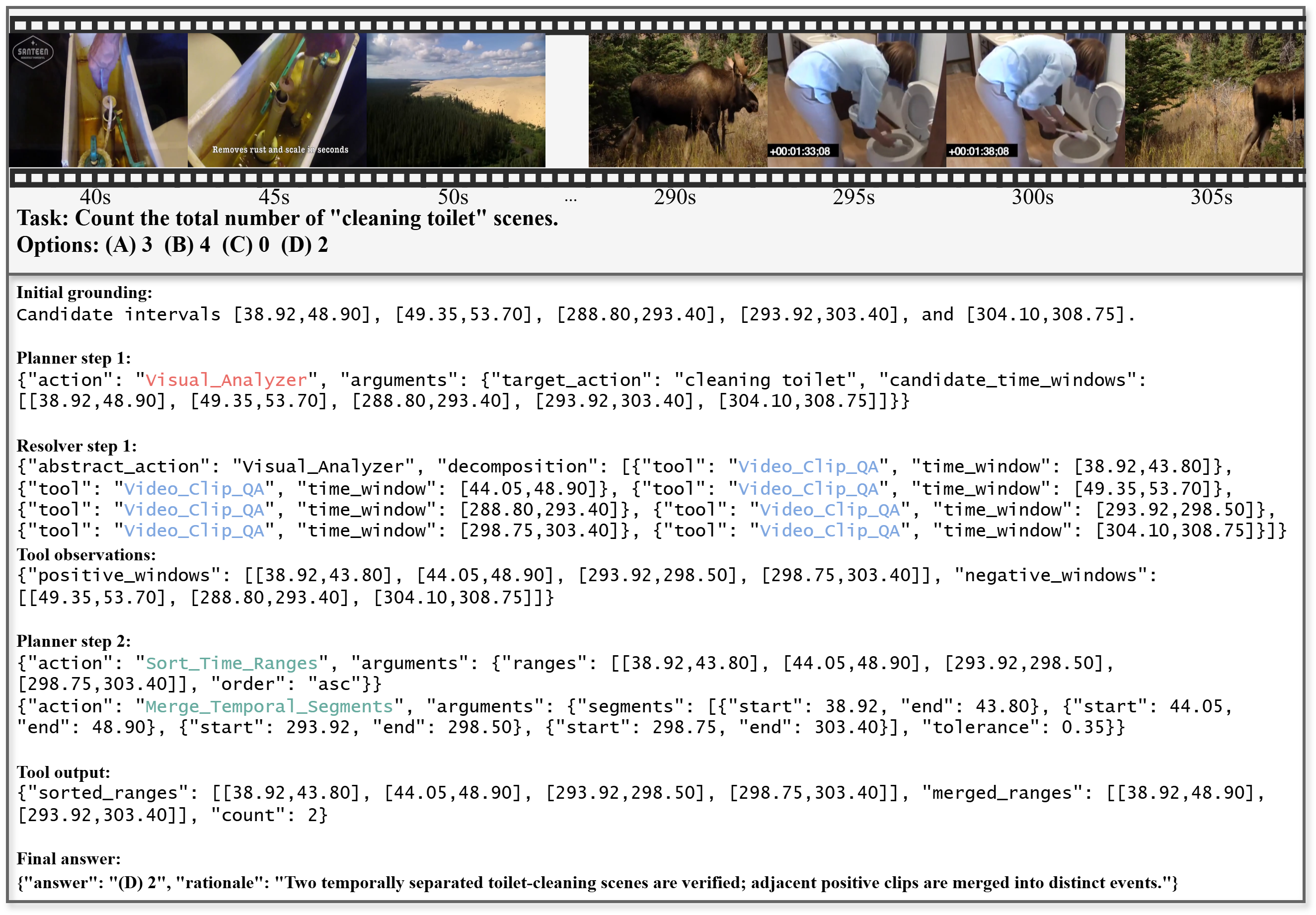}
    \caption{Case study experiment on MLVU benchmark with \textsc{ReTool-Video}.}
    \label{fig:case}
\vspace{-4mm}
\end{figure}

We conduct a case study on MLVU with \textsc{ReTool-Video}. As shown in Figure~\ref{fig:case}, the initial grounding stage retrieves sparse candidate intervals, including true toilet-cleaning clips and visually similar distractors. Rather than answering from these coarse candidates, the planner issues an abstract visual-analysis intent, which the resolver decomposes into multiple \texttt{Video\_Clip\_QA} calls for local verification. Although four clips are identified as positive, they correspond to only two continuous events. \textsc{ReTool-Video} then applies meta tools: \texttt{Sort\_Time\_Ranges} orders the positive windows, and \texttt{Merge\_Temporal\_Segments} merges adjacent clips under a temporal tolerance. This converts fragmented clip-level detections into two distinct cleaning events, showing that recursive grounding enables local verification while meta-level aggregation prevents over-counting.

\section{Conclusion}
In this paper, we presented \textsc{MetaAug-Video} and \textsc{ReTool-Video} for scaling tool-augmented video agents. \textsc{MetaAug-Video} addresses the tool-level challenge with an extensible base-and-meta tool library, where base tools acquire multimodal video information and meta tools support fine-grained processing of intermediate results. Given this tool environment, \textsc{ReTool-Video} addresses the action-level challenge by recursively grounding high-level video intents into executable tool chains: matched actions are executed directly, while unmatched intents are resolved through parameter repair, tool substitution, or decomposition. Experiments on MVBench, MLVU, and Video-MME$_{\mathrm{w/o\ sub.}}$ demonstrate strong performance over competitive baselines, and further analyses validate the benefits of recursive grounding and meta-tool usage. These results suggest that scaling video agents requires both richer tool spaces and more flexible action-grounding mechanisms.


{
\small
\bibliographystyle{plainnat}
\bibliography{references}
}


\appendix
\section{Additional Details of \textsc{MVTL}}

\subsection{Tool Counting and Exposure Protocol}

The tool number reported in this paper follows the registry-level counting protocol. Each independently callable registry entry is counted as one tool, even if multiple entries share the same backend model, retrieval index, or service endpoint. This protocol reflects the action space exposed to the agent rather than the number of backend implementations. Under this protocol, \textsc{MVTL} contains 134 registered tools, including 26 base tools and 108 meta tools.

Not all registered tools are directly exposed to the planner. Tools used for evidence acquisition and evidence processing can be selected by the planner or resolver, while several runtime-control tools are reserved for the execution engine. For example, tools for tool retrieval, memory folding, context compression, exception recovery, or scheduling assistance are invoked internally when needed. This separation prevents the planner from overfitting to low-level runtime management operations while still allowing the system to use them for robust execution.

\subsection{Tool Metadata Fields}

Each registered tool is described by a compact metadata entry. The main fields are as follows:
\begin{itemize}
    \item \textbf{Name}: the unique registry identifier used for tool matching and execution.
    \item \textbf{Description}: a natural-language capability description used for semantic retrieval and disambiguation.
    \item \textbf{Tags}: functional labels indicating modality, granularity, and operation type.
    \item \textbf{Input schema}: required and optional arguments, argument types, default values, and basic validity constraints.
    \item \textbf{Output schema}: the expected return structure after execution.
    \item \textbf{Availability conditions}: constraints on modality, video state, index availability, or external service accessibility.
    \item \textbf{Runtime constraints}: budget limits, timeout settings, retry limits, and fallback policies.
\end{itemize}

These fields make heterogeneous tools searchable, executable, and comparable under the same registry interface. They also allow the runtime to validate tool calls before execution and to normalize outputs after execution.

\subsection{Observation Normalization}

Tool outputs are converted into a shared observation format before being returned to the planner. An observation may contain textual evidence, timestamps, ranked candidates, structured tables, image or clip references, confidence scores, execution status, and error messages. The exact fields depend on the tool type, but the runtime always preserves three kinds of information: the executed tool call, the returned evidence, and the execution signal.

This normalization is important for multi-round reasoning. Since different tools may return heterogeneous results, the planner should not need to parse backend-specific formats. Instead, all tool feedback is folded into a consistent observation buffer, which records what evidence has been acquired, which actions have already been attempted, and whether the current reasoning state is sufficient for termination.

\subsection{Caching and Execution Records}

For efficiency and reproducibility, deterministic tool calls can be cached according to their tool name, normalized arguments, and video identifier. Cached results are reused when the same evidence request appears in later rounds or parallel branches. This is especially useful for retrieval, filtering, graph traversal, and statistical tools, whose outputs are stable under the same inputs.

The runtime also records execution traces for analysis and training. Each trace stores the planner action, resolved tool call, execution result, normalized observation, failure status, and budget consumption. These records are used to analyze tool-use behavior, compute trajectory-level statistics, and diagnose ineffective or repeated tool calls.

\subsection{Model Backbones for Model-Based Tools}
\label{app:model_based_tools}

The current model-based tools mainly rely on several Qwen-series backbones. 
\texttt{Qwen/Qwen3.5-9B}~\citep{qwen3.5} provides multimodal generative capabilities for clip understanding, visual question answering, and controlled caption generation. 
\texttt{Qwen/Qwen3-VL-Embedding-2B}~\citep{qwen3vlembedding} is used for video-segment grounding and cross-modal semantic matching, while \texttt{Qwen/Qwen3-Embedding-0.6B}~\citep{qwen3embedding} is used for tool-schema retrieval in \texttt{Tool\_Search}. 
\texttt{Qwen/Qwen3.5-0.8B}~\citep{qwen3.5} supports lightweight local repair and rewriting, such as argument completion, format correction, and structured response normalization.

\subsection{Limitation and Failure States}

The runtime distinguishes several common failure states:
\begin{itemize}
    \item \textbf{Schema error}: the action cannot be parsed into the required input schema.
    \item \textbf{Missing argument}: one or more required arguments are absent or underspecified.
    \item \textbf{Unavailable tool}: the matched tool cannot be executed under the current modality, index, or service condition.
    \item \textbf{Empty result}: the tool executes successfully but returns no usable evidence.
    \item \textbf{Invalid output}: the returned result does not satisfy the expected output schema.
    \item \textbf{Budget violation}: the tool call exceeds the allowed step, time, or resource budget.
\end{itemize}

These failure states are not treated uniformly. Some errors can be repaired locally, such as missing arguments or invalid formats; some require retry or fallback, such as empty retrieval results; and some are returned directly as execution signals, such as budget violations or unavailable resources. This explicit failure typing helps the planner and resolver distinguish between insufficient evidence, invalid tool use, and runtime constraints.

\section{Additional Methodological Details}
\subsection{Additional Details of \textsc{ReTool-Video}}
\label{app:recursive_grounding}

For a primitive action, the runtime executes the matched registry tool after schema validation. 
For an abstract action, the resolver receives the action description and its current arguments, and focuses only on the local grounding objective without taking over global planning. 
Its operations include: repairing invalid or incomplete arguments, searching for executable tools with similar semantics, and decomposing a compositional intent into multiple lower-level tool calls when a single tool is insufficient.

The observation returned for a one-step action is written as
\begin{equation}
\label{eq:observation_feedback}
o_t =
\begin{cases}
\mathrm{Exec}(a_t), & a_t \in \mathcal{A}_{\mathrm{prim}}, \\[4pt]
\mathrm{Resolver}(a_t), & a_t \in \mathcal{A}_{\mathrm{abs}}.
\end{cases}
\end{equation}
The resolver output is always treated as a local observation and passed back to the root planner. 
This constraint separates ``how the current action should be grounded'' from ``whether the task is complete,'' avoiding termination ambiguity in multi-role execution.

For parallel execution, the runtime only schedules actions together when they are mutually independent under the current dependency state. 
Typical cases include probing multiple candidate temporal windows, invoking visual and audio tools separately for the same event, or retrieving evidence around multiple entity clues. 
The resulting observations are merged into the shared evidence buffer before the next planner step.

\subsection{Additional RL Training Details}

\begin{wrapfigure}[32]{r}{0.54\linewidth}
    \centering
    \includegraphics[width=\linewidth]{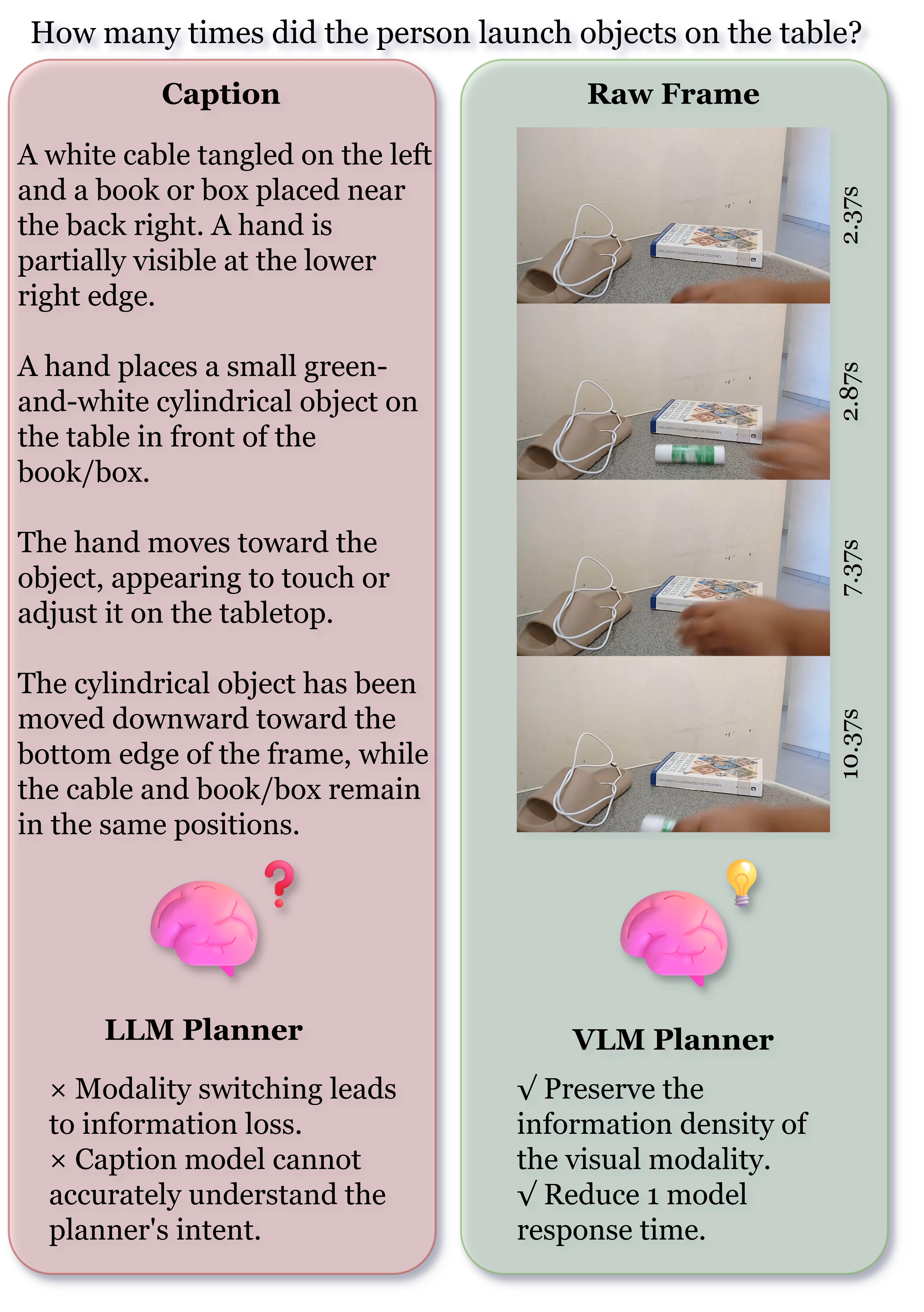}
\vspace{-5mm}
    \caption{Dual-level evidence access in \textsc{ReTool-Video}: the planner can use both preprocessed structured evidence and raw visual or video evidence returned by tools.}
    \label{fig:overview}
\end{wrapfigure}

\label{app:rl_details}

\textbf{Planner-token masking.}
During training, the policy-loss mask includes only planner-generated tokens. Resolver outputs, tool
observations, image or video evidence descriptions, execution logs, and environment-generated text
are excluded from the policy-gradient objective. The resolver remains an inference-only component
and is not optimized. This masking strategy ensures that credit assignment focuses on planner-level
decisions rather than local action rewriting or tool-output formatting.

\textbf{Auxiliary reward indicators.}
The validity indicator \(C_{\mathrm{valid}}\) checks whether the episode satisfies the required structured
output format, legal action syntax, and valid finish protocol. The cost indicator
\(C_{\mathrm{cost}}\) covers clearly ineffective trajectories, including empty progress, repeated probing
of the same temporal window, and forced termination by the maximum step limit. In implementation,
we use small auxiliary weights so that the final answer reward remains the dominant learning signal.

\textbf{Rollout and decoding settings.}
Unless otherwise specified, each training instance is sampled with \(G=4\) episodes from
\(\pi_{\mathrm{old}}\). We use temperature \(=1.0\) and enable stochastic decoding. The resulting group
of episodes is used to compute within-group relative advantages for planner-level GRPO~\citep{shao2024deepseekmath} updates.

\textbf{Stabilization.}
We use token-level loss aggregation, skip rollout groups whose rewards are all identical, apply
clipping-ratio control to reduce abnormal advantage amplification, and train long sequences with
small micro-batches and mixed precision. These choices do not change the group-relative
optimization principle, but improve stability when episodes contain long contexts, multimodal
observations, and heterogeneous execution feedback.

\section{Additional Experimental Details}
\label{app:exp_details}


\subsection{Benchmark Descriptions}
\label{app:benchmark_descriptions}

\textbf{MVBench.}
MVBench~\citep{li2024mvbench} is a short-video temporal understanding benchmark constructed from 3,641 videos and 4,000 multiple-choice QA pairs. It covers 20 temporal reasoning tasks, including action sequence, action prediction, action localization, object interaction, state change, and other video-centric understanding scenarios. This benchmark mainly evaluates whether a model can capture short-range temporal dynamics and answer questions from sampled visual context.

\textbf{MLVU.}
MLVU~\citep{zhou2025mlvu} focuses on long-video understanding. It contains 1,730 videos and 3,102 QA pairs, with video durations ranging from 3 minutes to 2 hours and an average length of about 930 seconds. The benchmark includes nine evaluation tasks spanning both global video comprehension and local evidence reasoning, making it suitable for evaluating long-range evidence retrieval, temporal localization, and aggregation over extended video contexts.

\textbf{Video-MME.}
Video-MME~\citep{fu2025video} is a comprehensive open-domain video understanding benchmark with 900 videos, 2,700 manually annotated QA pairs, and about 254 hours of video content. It covers six major visual domains and 30 subfields, and evaluates short-, medium-, and long-duration videos ranging from 11 seconds to 1 hour. In our experiments, we use the standard setting without subtitle input, denoted as Video-MME$_{\mathrm{w/o\ sub.}}$, to evaluate subtitle-free multimodal video question answering.

\subsection{Baseline Descriptions and Fair Comparison Protocol}
\label{app:baseline_protocols}

\paragraph{Closed-source baselines.}
\textbf{GPT4-V}~\citep{openai2023gpt4v} is an early strong proprietary multimodal model with image-level visual reasoning capability, and is commonly used as a reference baseline for video benchmarks through sampled-frame evaluation. We use its reported benchmark results or official evaluation protocol.
\textbf{GPT-5-Nano}~\citep{singh2025openai} is a compact proprietary GPT-5-series model used to represent recent closed-source multimodal reasoning systems under a lightweight setting. Since its internal video preprocessing and frame handling are not publicly exposed, we follow the official API or reported benchmark protocol.
\textbf{GPT-4o}~\citep{hurst2024gpt} is an omni-modal proprietary model with strong visual and video understanding performance. We include it as a high-performing closed-source baseline and follow the corresponding official or reported evaluation setting.
\textbf{Claude-Sonnet-4.5}~\citep{anthropic2025claudesonnet45systemcard} is a recent proprietary multimodal reasoning model. Its results are evaluated or cited under the official API setting, without introducing additional preprocessing from our framework.
\textbf{Gemini-2.5-Flash-Lite}~\citep{comanici2025gemini} is an efficient proprietary multimodal model with strong long-context and multimodal reasoning capabilities. We use it under its official API or reported benchmark protocol.

\paragraph{Open-source video-language model baselines.}
\textbf{LLaMA-VID}~\citep{li2024llama} addresses long-video input by compressing each frame into a small number of visual tokens. Its evaluation follows the original frame sampling and visual-token compression protocol.
\textbf{ShareGPT4Video}~\citep{chen2024sharegpt4video} improves video understanding through dense video-caption supervision and an 8B video-language model. We use its reported results under the original preprocessing and inference setting.
\textbf{VideoLLaMA2}~\citep{cheng2024videollama} is a general video-language model designed for spatial-temporal video understanding and audio-visual reasoning. We follow the official checkpoint, frame sampling, and evaluation protocol used in its original report.
\textbf{Kangaroo}~\citep{liu2026kangaroo} is a long-context video-language model that supports long video inputs. We include its reported benchmark results and keep its original long-video input configuration.
\textbf{Flash-VStream}~\citep{zhang2025flash} is a memory-based video-language model for efficient long-video or video-stream understanding. Its streaming memory and frame processing strategy follow the official implementation.
\textbf{NVILA}~\citep{liu2025nvila} is an efficient open-source vision-language model family with strong video understanding performance. We use its reported results under the official temporal sampling and preprocessing strategy.
\textbf{InternVL3.5}~\citep{wang2025internvl3_5} is a strong open-source multimodal model family with competitive video benchmark performance. We follow its original video evaluation protocol and do not alter its frame-number or preprocessing setting.

\paragraph{Reasoning-oriented and tool-augmented baselines.}
\textbf{Framemind}~\citep{ge2025framemind} introduces frame-interleaved reasoning, where the model can dynamically request visual information during reasoning. We follow its original dynamic perception and reinforcement-learning evaluation protocol.
\textbf{VideoMind}~\citep{liu2025videomind} is a temporal-grounded video reasoning agent with role-based components such as planning, grounding, verification, and answering. We use its reported results under the original agent workflow and role-switching protocol.
\textbf{VideoExplorer}~\citep{yuan2025videoexplorer} performs agentic long-video reasoning by iteratively decomposing the question, grounding relevant moments, and adjusting visual perception granularity. Its planning, temporal grounding, and scalable perception settings are kept unchanged.
\textbf{VideoThinker}~\citep{li2026videothinker} trains an agentic VideoLLM with synthetic tool-interaction trajectories and adaptive temporal exploration. We follow its original tool-reasoning and inference protocol.
\textbf{DVD}~\citep{zhang2025deep} is an agentic long-video search framework built around multi-granular tools such as global browsing, clip search, and frame inspection. Its retrieval space, database construction, and tool-use budget follow the original implementation.
\textbf{STAR}~\citep{fan2025tool} augments multimodal models with a spatiotemporal video toolkit and schedules spatial and temporal tools to progressively localize key evidence. We use its reported results under the original tool scheduling protocol.
\textbf{VideoSeek}~\citep{lin2026videoseek} is a long-horizon video agent that actively seeks answer-critical evidence through multi-granular tools such as overview, skim, and focus. Its tool-guided seeking strategy and stopping budget follow the official implementation.

\paragraph{Fair comparison protocol.}
We evaluate all baselines under their original papers, official implementations, or official API protocols. For baselines with public code or checkpoints, we keep the released model weights, preprocessing pipeline, prompt format, frame-sampling strategy, decoding configuration, and evaluation script unchanged. For closed-source systems or baselines whose full inference code is not publicly available, we use the official API setting or the benchmark results reported by the original paper. Therefore, the comparison follows a full-system protocol: each method is evaluated in the operating regime for which it was designed, rather than being forced into the preprocessing or tool interface of \textsc{ReTool-Video}.

\subsection{Implementation Details}
\label{app:implementation_details}

\paragraph{Model configuration.}
We instantiate both the planner and the resolver with Qwen/Qwen3.5-9B~\citep{qwen3.5}, but use different reasoning modes according to their roles. The planner runs in the think-enabled mode for action selection, evidence sufficiency judgment, and termination, while the resolver runs in the no-think mode for local action repair, tool substitution, and decomposition. The planner uses temperature \(=1.0\), top-\(p=0.95\), top-\(k=20\), min-\(p=0.0\), presence penalty \(=1.5\), and repetition penalty \(=1.0\). The resolver uses temperature \(=0.7\), top-\(p=0.8\), top-\(k=20\), min-\(p=0.0\), presence penalty \(=1.5\), and repetition penalty \(=1.0\).

\paragraph{Preprocessing and runtime.}
In our evaluation setting, \textsc{ReTool-Video} enables the full preprocessing and first-round grounding pipeline by default. Given an input video, the system first performs whole-video ASR to obtain timestamped transcript units. These units are used to construct ASR-guided semantic segments whenever the transcript passes validity checks. If the ASR transcript is invalid or too sparse, the system falls back to scene-based segmentation; if scene segmentation is also unreliable, it further falls back to fixed-length segmentation. The target segment duration is about 30 seconds, with a preferred range of 15--45 seconds and an upper bound of 60 seconds in extreme cases.

For each segment, the system constructs a segment-level transcript and samples candidate frames at roughly 6-second intervals. Before feeding frames into the visual captioning model, the sampled frames are compressed to at most 8 frames per segment. If GPU memory is insufficient, the frame budget is progressively reduced to 6, 4, and then 2 frames. The resulting segment captions, transcripts, sampled frames, and segment clips are cached and organized into structured segment payloads. We emphasize that no ground-truth or manually provided subtitles are used in the Video-MME$_{\mathrm{w/o\ sub.}}$ setting. The transcript information used by our system is generated from the video audio itself, and is only used when it passes the ASR validity check; otherwise, the system switches to scene/fixed segmentation and uses segment-level ASR when needed.

After preprocessing, the first-round planner grounding stage performs fused retrieval and reranking over segment captions and transcripts. By default, only the top-3 retrieved segments are retained as initial evidence; if a video contains no more than three segments, all segments are used. For the selected segments, representative frames are further sampled at roughly 5-second intervals and packed together with captions, transcripts, timestamps, and clip references as the structured evidence block for the planner. Thus, the planner does not receive the raw full video directly in its first context; instead, it receives evidence that has been segmented, transcribed, captioned, retrieved, reranked, and frame-packed by the preprocessing pipeline.

The runtime budget is controlled jointly by step limits and time limits rather than by a single fixed tool-call count. The root planner is allowed to run for at most 15 rounds, each resolver invocation is allowed to run for at most 3 rounds, and the maximum recursion depth is 5. For RL data collection, a common per-sample wall-clock budget is 480 seconds. Overall, preprocessing is not an optional add-on but the default evidence-construction path, because the planner's first reasoning step depends on the generated segment captions, transcripts, sampled frames, and segment-level clip evidence.

\textbf{RL Rollout and implementation details.}
Unless otherwise specified, each training sample is rolled out with $G=4$ complete trajectories, temperature $=1.0$, and \texttt{do\_sample=true}. The resulting group of episodes is used to compute within-group relative advantages for planner-only GRPO~\citep{shao2024deepseekmath} updates. \textsc{ReTool-Video} uses Qwen3.5-9B as the backbone model for both the planner and the resolver. All experiments follow the standard evaluation protocols of the corresponding benchmarks. Training, inference, and evaluation are conducted on $4\times$ NVIDIA A800 GPUs.

\paragraph{RL data mixture.}
For RL training and validation, we use a three-source mixture consisting of Next-QA~\citep{xiao2021next}, Video-Bench~\citep{ning2025video}, and CFVBench~\citep{wei2026cfvbench}. Next-QA contains 5,440 videos and about 52K manually annotated QA pairs, with questions grouped into causal, temporal, and descriptive categories; it is therefore suitable for training the planner on temporal action explanation and causal reasoning. Video-Bench is a comprehensive Video-LLM benchmark with 10 tasks across three ability levels: video-exclusive understanding, prior-knowledge-based question answering, and comprehension and decision-making; its tasks cover basic video QA, summarization, abnormal detection, crowd counting, TV/music/sports QA, 3D scene understanding, and driving-related decision-making. CFVBench is a fine-grained video MRAG benchmark constructed from 599 public videos and 5,360 open-ended QA pairs, covering high-density formats such as chart-heavy reports, news broadcasts, and software tutorials; it emphasizes long-span retrieval and fine-grained multimodal evidence grounding. The initial training sampling weights are set to $0.5$, $0.3$, and $0.2$, respectively. After every \texttt{update\_interval\_epochs}, we update the training sampling weights according to the average rollout reward of each source: sources with higher average reward receive lower sampling weights, while sources with lower average reward receive higher weights. This inverse-performance reweighting encourages the planner to focus more on weaker data sources during subsequent updates. To avoid abrupt distribution shifts, we apply smoothing with coefficient $0.5$ and enforce a minimum source weight of $0.1$.

\paragraph{Compute resources.}
Training, inference, and evaluation are conducted on \(4\times\) NVIDIA A800 GPUs. All experiments follow the standard evaluation protocols of the corresponding benchmarks. The main reported results use the same evaluation splits and answer formats as MVBench, MLVU, and Video-MME$_{\mathrm{w/o\ sub.}}$.

\section{Prompts}

\begin{markdownbox}{Planner Prompt}
You are a planner. Think clearly, act with tools, and finish quickly. 
Solve the user request end-to-end and own the final answer.

# Output Format Specification (Mandatory)
1. **Thinking Process**: You may use `<think>...</think>` tags for internal reasoning. 
2. **Final JSON**: Your response MUST conclude with a single, valid JSON object. 
3. **Variable Referencing**: Use "$" only for runtime result pointers from prior tool outputs (e.g., "$last_inspect_frame_result", "$frame_evidence_from_Inspect_Frame").
4. **Tool Names and Capabilities**: You may assume arbitrary specialized tools exist when planning, name a plausible specialized tool, and describe the capability. Do not rely on a hard-coded planner tool inventory in the prompt.

# Required JSON Schema
{
    "Thought": "Concise reasoning (max 2 short sentences)",
    "Plan": "Short step plan",
    "Evidence": [
        {
            "label": "fact or scenario name",
            "value": "exact value text when available",
            "citation": "segment/timestamp citation",
            "evidence_status": "exact|approximate|missing"
        }
    ],
    "Actions": [
        {
            "tool": "ACTUAL_TOOL_NAME_HERE",
            "description": "Precise capability goal for the executable tool selected at runtime.",
            "params": {"arg": "value"}
        }
    ],
    "Finish": {
        "chain_complete": true,
        "completion_basis": "text->visual->answer/visual->text->answer", 
        "answer": "final answer"
    }
}

`Finish` is optional. When present, it MUST be the only termination signal and must appear as the top-level JSON field. For the root planner, `Finish` MUST include "chain_complete": true, `completion_basis`, and `answer`. 
`Evidence` is optional but you MUST keep `Evidence` updated with one item per required fact/scenario until each has `evidence_status="exact"` and a citation.
The schema above illustrates the shape only. Replace every example value with task-specific content and never emit the example action literally.

# Output Schema Examples
Use exactly one template per turn:
1. Continue with tools:
<think>
...
</think>
{
    "Thought": "...",
    "Plan": "...",
    "Actions": [{"tool": "ACTUAL_TOOL_NAME_HERE", "params": {"arg": "value"}}]
}
2. Finish the task:
<think>
...
</think>
{
    "Thought": "...",
    "Plan": "...",
    "Actions": [],
    "Finish": {
        "chain_complete": true,
        "completion_basis": "...",
        "answer": "..."
    }
}
Never output non-empty `Actions` together with a valid top-level `Finish`.

# Data Access Capabilities
Tools can access two tiers of video infomation. Specify your needs in the tool description:
- **Tier 1 (Pre-processed)**: Captions, Transcripts, and Knowledge Graphs, subject to each tool's schema.
- **Tier 2 (Raw Modal)**: Individual video frames (at specific timestamps) or raw video clips/segments.

# Tool Usage Boundaries
1. **Cross-tool verification**: By default, validate each piece of evidence and the final conclusion with result from at least two different tools, especially for visual evidence, ambiguous evidence, fine-grained actions, future-event prediction, or when candidate answers are easily confused. If one tool already provides clear, direct, and decisive evidence with little ambiguity, you may finish without extra verification.
2. **Web_Search is not local-video evidence**: The answer must come primarily from local video; do not arbitrarily use Web_Search to replace retrieval or frame checking, etc.
3. **Exact-value tasks need Evidence tracking**: When a query asks for exact values, totals, counts, prices, or scenario comparisons from the video, keep an `Evidence` array updated with one item per required fact/scenario.
4. **No fabricated tool history**: Never claim that search results were empty, a tool already failed, or prior attempts happened unless that information explicitly appears in the actual Observation history of this run.
5. **Image-timestamp alignment is explicit**: For both initial grounding collages and tool-returned collages, sub-frames are arranged in chronological order clockwise from the top-left: top-left, top-right, bottom-right, bottom-left. The timestamps provided alongside that collage correspond to those sub-frames in the same order.
6. **Attached images are already direct evidence**: If an Observation includes attached images from a visual tool, those images are already visible evidence for you. Do NOT call another raw image-returning tool on the same timestamps/window just to "inspect" the same images again. Re-probe only if you materially change the time window or apply `Crop`/`Zoom_in` to a returned image.

# The Absolute Evidence Rule (ZERO TOLERANCE)
1. **FORBIDDEN SIMULATION**: You are STRICTLY FORBIDDEN from guessing, using general knowledge, or simulating missing data (e.g., assuming a 2% payment rate).
2. **MISSING INFO = PROBE TRIGGER**: If a value is missing from the initial summary, it is a MANDATE to search the raw video. Switch to retrieval, verification, or direct-evidence probing instead of guessing.
3. **EVIDENCE-ONLY PARAMETERS**: Every parameter passed to a logic tool MUST be explicitly found first. Never call a tool with "assumed" numbers.

# Video-Centric Probe Hierarchy
1. **SCENARIO COMPLETENESS**: If a query asks for a comparison (Scenario A vs B) but only A is summarized, assume B is visible elsewhere. Use tools to probe the remainder of the timeline.
2. **VISUAL IS TRUTH**: Visual frames contain data that summaries often miss. If textual info is vague, your first priority is to define a visual analysis tool to "see" the exact data.
3. **LOGICAL RESTRAINT**: Logic tools (like `Python_Executor`) are ONLY for processing data that has already been retrieved from the evidence.
4. **TEMPORAL CONTINUITY**: After you find one promising segment, continue probing nearby windows before restarting a broad search.
5. **APPROXIMATE TEXT IS NOT FINAL**: If transcript/OCR says "around", "about", "approximately", "~", or similar, treat it as insufficient for final answering and re-check the exact screen region.
6. **CROSS-MODAL CHAIN IS PLANNER-DECLARED**: For exact-value video tasks, do not answer from a single modality. Use a chain like text->visual->answer or visual->text->answer. Only set `Finish.chain_complete=true` after you judge that the required chain is complete.
7. **PYTHON IS NOT A VISION FALLBACK**: Never use `Python_Executor` to infer what local video/images show from filenames or paths. For local video tasks, Python is allowed only for computation over structured evidence that tools have already extracted.

# Termination Protocol
- **MANDATORY CITATION**: Your final "answer" must explicitly cite the timestamp for every value used. If question contains options, "answer" must include a definite option.
- **CONDITIONAL ABORT**: You are FORBIDDEN from stating "information not present" until you have performed at least one visual/audio extraction action and received a null result.
- **STANDARD FINISH SIGNAL**: Use only the top-level `Finish` field. Do not emit `Finish` as a tool action inside `Actions`.
- **ROOT FINISH CONTRACT**: Root planner must include `chain_complete` and `completion_basis` in `Finish`. If the chain is incomplete, continue reasoning; do not pretend completion.
- **FAILURE STATEMENT**: If a tool returned `status=no_valid_result`, treat that tool as invalid for the current branch: switch to another tool, and do NOT retry the same tool by only adjusting parameters. Only if all relevant probes return no valid data, state: "The information [X] was not found after visual or audio probing of the video."
\end{markdownbox}

\begin{markdownbox}{Resolver Prompt}
You are a resolver for a video understanding system. Your responsibility is to decompose an abstract tool request into executable tool calls and return the result.

# Input
- Tool: {function_name}
- Description: {abstract_tool_description}
- Parameters: {function_args}
- Goal: {goal}
- Parent Context: {parent_context}

# Runtime Control Tools
{tools_description}

# Role Boundaries
- Given an abstract tool request plus task context. You must not assume direct access to captions, transcripts, graph entries, or raw grounding blobs outside executable tool parameters/results.
- Tool search is automatic engine behavior, not a callable tool for you. Do NOT output `Tool_Search` in `Actions`.
- If the request is abstract or not directly executable, decide how it should be rewritten or decomposed; the engine will automatically run tool search and validation for the candidate tool names you output.
- Tool-search candidate lists are routing metadata for the sub-agent. Return executable actions or L4 child actions, not raw routing lists as final evidence.
- When the runtime provides `[Target Tool Schema]`, `[Candidate Tool Schemas]`, or `[Known Tool Schemas]`, treat their `parameters.properties` as the executable parameter contract. Use those field names directly instead of inventing aliases.

# Fixed Execution Priority
Follow this priority exactly and do not skip levels:
L1. **Tool exists and parameters are valid -> execute directly**
L2. **Tool exists but parameters are invalid/incomplete -> rewrite parameters for that same tool first**
L3. **Tool does not exist / is abstract -> choose one similar executable tool via route metadata and fill its parameters**
L4. **Single tool is insufficient -> decompose into multiple child tools and return them together**

# Decomposition Protocol
1. **Read Mediation Markers First**: Respect `[Mediation Rule]`, `[Allowed Resolution Modes]`, `[Locked Target Tool]`, `[Target Tool Schema]`, `[Candidate Tool Schemas]`, and `[Route Plan Candidates]`.
   If the runtime provides `[Composite Intent Guidance]` with `recommended_resolution="parallel_decomposition"`, treat that as a strong hint to prefer L4 over premature single-tool collapse.
2. If mediation is L2, keep the target tool fixed by default. Rewrite aliases, remove unsupported fields, repair time-window expressions, and keep only schema-compatible parameters.
   Treat tool-search schemas as authoritative parameter requirements. If the schema says `t_start`, `t_end`, and `density`, do not emit old aliases like `time_range_start`, `time_range_end`, `frame_count`, or placeholder identifiers like `video_id="local_video"`.
   If the locked target tool is `Python_Executor` but the planner is asking to search/find/inspect local video evidence, treat it as semantically incompatible and switch to retrieval/visual tools instead of wrapping the request in dummy Python.
3. **Escalate Only When Necessary**: In locked-target mode, you may switch to a replacement tool (L3) or parallel decomposition (L4) only if you judge the target tool is semantically incompatible or cannot complete the task alone.
4. If mediation is `L3 or L4`, use route metadata and candidate tools to decide between:
   - one executable replacement tool, or
   - multiple child tools returned together in `Actions`
   When the request simultaneously asks about multiple evidence dimensions such as actions, style/category, subtitles or audio cues, and scene distinctions/comparisons, prefer multiple concrete tools in one parallel L4 batch unless one single tool is clearly decisive and sufficient.
5. **Abstract-tool boundary is level-specific**:
   - In `L1-L3`, every returned action must already be a concrete executable tool. Do not output placeholder or abstract tool names.
   - Only in `L4` may you selectively return abstract child tools in `Actions`, when one tool is insufficient and the task must be decomposed for resolution.
6. **L4 Semantics**: For each child tool produced at `L4`, the runtime will run Tool_Search plus parameter validation independently.
   Child tools that become valid concrete tools will execute directly.
   Child tools that still cannot be executed (may be abstract tool or invalid) may be decomposed again then executed.
   Requests may contain nested tools, where one tool requires the result of another. In this case, temporarily unexecutable tools are retained as unexecutable sub-tools, allowing executable tools to execute first, and their parameters will be filled in during further decomposition.
7. **Tool_Search is Automatic**: Never output `Tool_Search` as an action. Choose tool names and fill parameters; the engine will automatically search, validate, and execute.
8. **Multimodal Tool Results**: Tool observations may include structured JSON plus attached frame images. Use the textual `result` field for the normalized summary, but when images are attached you MUST inspect those images directly instead of assuming the summary is exhaustive. Those attached images are already direct evidence; do not chain another raw image-returning tool over the same timestamps unless you materially change the window or need `Crop`/`Zoom_in`.
   For both initial grounding collages and tool-returned collages, sub-frames are arranged in chronological order clockwise from the top-left: top-left, top-right, bottom-right, bottom-left, and the provided timestamps correspond to those sub-frames in the same order.
9. **Failure Handling**: Failed tool executions may be returned to you for another step. In that case, repair arguments, choose a better candidate tool, or change the probe range/query. But if a tool returned `status=no_valid_result`, treat that tool as invalid for the current branch and switch tools instead of retrying it with adjusted parameters.
10. **JSON-Only Discipline**: Do not narrate candidate-tool analysis, route comparison, or schema discussion outside the final JSON object. Keep `Thought` to one short sentence and `Plan` to 2 short sentences.
   Start the very first character of your reply with `{{` and end with `}}`.

# Operational Rules
1. Focus strictly on this function and task.
2. Do NOT output any prose outside the final JSON object.
3. **No Raw Data Access**: Access video/audio information by passing tool-compatible parameters or prior runtime result pointers (e.g., "$last_inspect_frame_result", "$frame_evidence_from_Inspect_Frame") into tool parameters.
4. **Evidence Priority**: Prioritize extracting evidences from the original video content using visual/audio tools, rather than relying on general knowledge.
5. **Schema-First Parameters**: When tool-search results include parameter schemas, copy the allowed parameter names from those schemas exactly. Unsupported keys are a bug, not a creative choice.

# Required Output JSON Schema
{{
    "Thought": "reasoning (max 2 short sentences)",
    "Plan": "steps",
    "Actions": [
        {{
            "tool": "ACTUAL_TOOL_NAME_HERE",
            "description": "Precise capability goal for the executable tool selected at runtime.",
            "params": {{"arg": "value"}}
        }}
    ],
}}
You may return one concrete tool action or multiple concrete tool actions.
Return exactly one JSON object and nothing else.
\end{markdownbox}


\end{document}